\documentclass[conference]{IEEEtran}

\IEEEoverridecommandlockouts
\usepackage{cite}
\usepackage{amsmath,amssymb,amsfonts}
\usepackage{algorithmic}
\usepackage{graphicx}
\usepackage{textcomp}
\usepackage{xcolor}

\usepackage{times}
\usepackage[T1]{fontenc}
\usepackage{url}
\usepackage{amsthm}
\usepackage{booktabs}
\usepackage{algorithm}
\usepackage{makecell}
\usepackage[most]{tcolorbox}
\usepackage{rotating}  
\usepackage{array}
\usepackage{caption}
\usepackage{multirow}

\newcommand{\ModelName}{{MuPaS}}
\newcommand{\vanillaFT}{{VanillaSFT}}
\newcommand{\MultiParty}{{MPD}}

\def\BibTeX{{\rm B\kern-.05em{\sc i\kern-.025em b}\kern-.08em
    T\kern-.1667em\lower.7ex\hbox{E}\kern-.125emX}}
\begin{document}

\title{Multi-Party Supervised Fine-tuning of Language Models for Multi-Party Dialogue Generation
}




\author{
\IEEEauthorblockN{
Xiaoyu Wang\IEEEauthorrefmark{1,*},
Ningyuan Xi\IEEEauthorrefmark{2,*},
Teng Chen\IEEEauthorrefmark{3},
Qingqing Gu\IEEEauthorrefmark{3},
Yue Zhao\IEEEauthorrefmark{3},
Xiaokai Chen\IEEEauthorrefmark{1},\\
Zhonglin Jiang\IEEEauthorrefmark{3},
Yong Chen\IEEEauthorrefmark{3},
Luo Ji\IEEEauthorrefmark{3,†}
}

\vspace{0.2cm}
\IEEEauthorblockA{\IEEEauthorrefmark{1}Beijing Institute of Technology, Beijing, China}
\IEEEauthorblockA{\IEEEauthorrefmark{2}Beihang University, Beijing, China}
\IEEEauthorblockA{\IEEEauthorrefmark{3}Geely AI Lab, Beijing, China}

3220230388@bit.edu.cn, 21373102@buaa.edu.cn, \{Teng.Chen2, Qingqing.Gu3, Yue.Zhao17\}@Geely.com,\\ chenxiaokai@bit.edu.cn,\{zhonglin.jiang, yong.chen, Luo.Ji1\}@Geely.com
}


\maketitle

\renewcommand{\thefootnote}{\fnsymbol{footnote}}
\footnotetext[1]{The first two authors contributed equally to this research; work was done during their internship at Geely.}
\footnotetext[2]{Corresponding author: Luo.Ji1@Geely.com}
\renewcommand{\thefootnote}{\arabic{footnote}}

\begin{abstract}
Large Language Models (LLM) are usually fine-tuned to participate in dyadic or two-party dialogues, which can not adapt well to multi-party dialogues (MPD), which hinders their applications in such scenarios including multi-personal meetings, discussions and daily communication. Previous LLM-based researches mainly focus on the multi-agent framework, while their base LLMs are still pairwisely fine-tuned. In this work, we design a multi-party fine-tuning framework (MuPaS) for LLMs on the multi-party dialogue datasets, and prove such a straightforward framework can let the LLM align with the multi-party conversation style efficiently and effectively. We also design two training strategies which can convert MuPaS into the MPD simulator. Substantial experiments show that MuPaS can achieve state-of-the-art multi-party response, higher accuracy of the-next-speaker prediction, higher human and automatic evaluated utterance qualities, and can even generate reasonably with out-of-distribution scene, topic and role descriptions. The MuPaS framework bridges the LLM training with more complicated multi-party applications, such as conversation generation, virtual rehearsal or meta-universe.
\end{abstract}


\section{Introduction}

\begin{figure}[t]
  \includegraphics[width=1\linewidth]{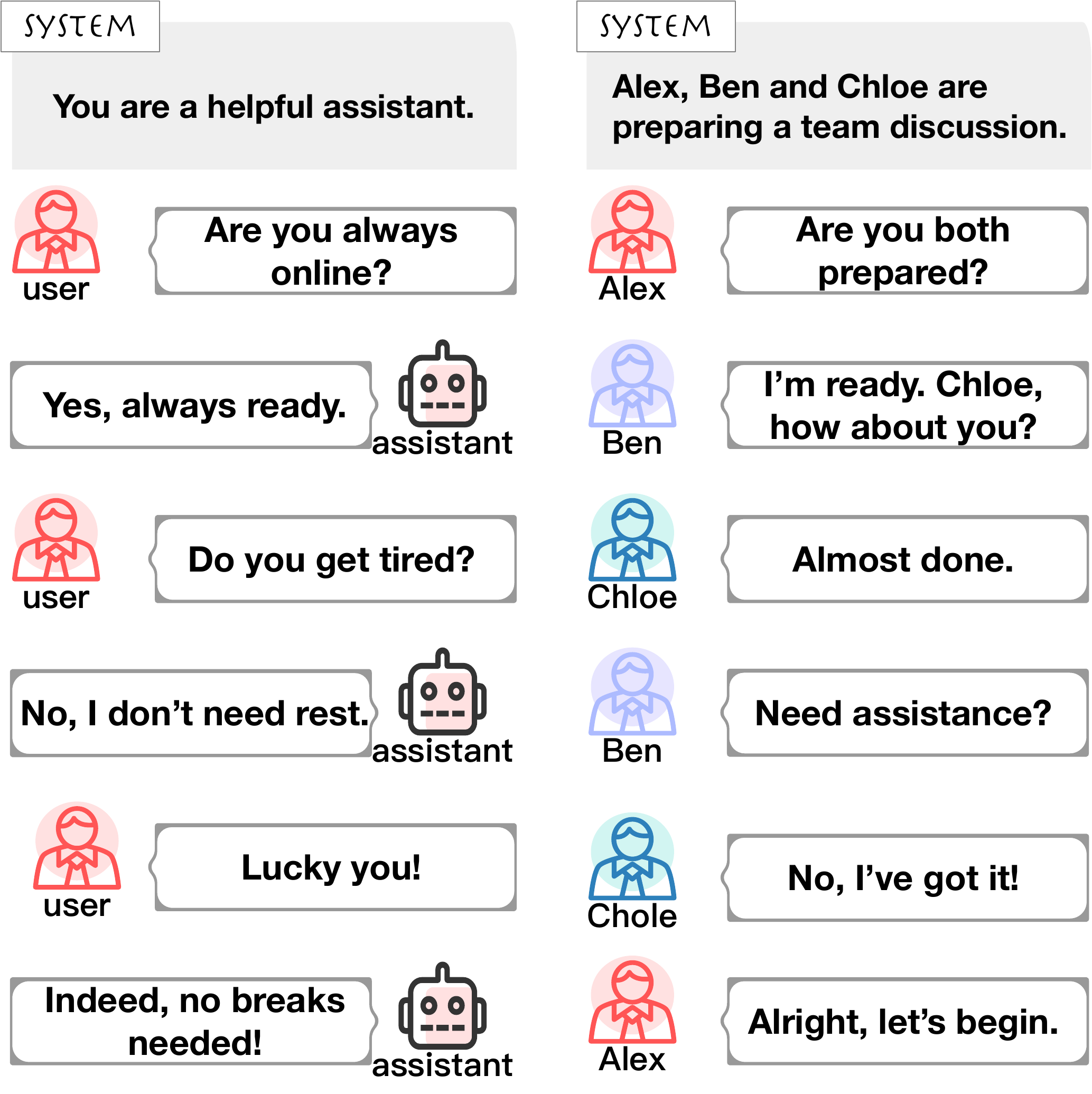} 
  \caption {The paradigm shift from the conventional two-party dialogue (Left) to the multi-party dialogue (Right). The training target also change from the helpful assistant to different possible persona or roles.}
  \label{fig:paradigm}
\end{figure}

In recent years, large language models (LLM) have demonstrated significant advancements in dyadic conversational contexts, such as question-answering systems and chatbot companions. Such applications are primarily structured around binary dialogue attendants (typically `human' and `assistant'), which are supported by widespread open-source models and datasets. However, many real-world scenarios instead encompass Multi-Party Dialogues (MPD) (Some papers instead name this scenario by multi-party conversation (MPC).), such as team meetings, classroom discussions, court or academic debates, or simply daily conversions with multiple humans involved \cite{mahajanNeedThoughtfulData2021, ganeshSurveyChallengesMethods2023}. Instead of responding to a single user's query, in such a case, the dialog system needs to understand conversation contexts from multiple users, determine whether to speak or not, and reasonably participate in potential multiple concurrent topics. Novel modeling technique is therefore required to adapt to this different dialogue paradigm.

Previous researches have sought to address the unique challenges of {\MultiParty} modeling, such as MIDS  \cite{yangEndEndPersonalizedHumorous2019}, ChatMDG \cite{liChatMDGDiscourseParsing2024}, ReDE \cite{shenGenericDependencyModeling2023}, SDMPED \cite{zhuMultiPartyEmpatheticDialogue2022} and MPC-BERT \cite{guMPCBERTPreTrainedLanguage2021}. However, these works are mostly RNN, Bert or Graph-based, which have not yet leveraged the semantic knowledge and generation capabilities of modern LLM, and is difficult to scale up and generalize to different domains. There are also LLM-based {\MultiParty} approaches, which are generally based on multi-agent systems. Each LLM agent still adheres to a binary interaction framework, and the conversations might happen between different pairs of agents which do not happen simultaneously and concurrently. Besides that, conventional LLMs are fine-tuned with human-assistant dialogues with alignment with the helpful, impartial, and conservative conversation style. In short, there is still not a single LLM-based, purely training framework which allows the model to learn from {\MultiParty} directly, unify the response generation and the speaker in a uniform manner, and portray different persona styles (either by data-driven or system prompted).





In this work, we propose a \textbf{Mu}lti-\textbf{Pa}rty \textbf{S}upervised (\textbf{\ModelName}) fine-tuning framework to train LLMs as the {\MultiParty} participants. Starting from a conventional instruct version of LLM which can handle two-party conversations, we provide an extra post-training stage in which the {\MultiParty} datasets are supervised fine-tuned, such that adapt its chat capability from the two-party to the multi-party format. As indicated by Figure \ref{fig:paradigm}, we preprocess the dataset by annotating lists of roles and sample-wise scene descriptions. We allow the LLM to be fine-tuned with each role's utterance while other roles are masked as context. We further apply this approach as the basis of {\MultiParty} builder by designing the model to recognize the next speaker simultaneously. By thoroughly designed experiments, we find our {\ModelName} can both generate state-of-the-art response quality and achieve the highest next-speaker prediction accuracy, compared with previous baselines, within the {\MultiParty} scope. We also provide several entertainment {\MultiParty} case simulations which indicate our approach can generate stylized and dramatic scripts. Our study shed some light on the constructions of AI-involved discussion or debate, and multi-agent environments. Our main contributions are as follows:
\begin{itemize}
    \item We propose a purely training-based approach to let the LLM participate in multi-party dialogue.
    \item We develop two strategies to build a multi-party dialogue simulator, which could be applied to show-script creation, scenario simulation, or debate rehearsal.
    \item We design experiments to verify the effectiveness of our methodology, including the next-speaker prediction, and assessment of multi-party response qualities.
    \item Our approach can be further employed as the multi-party dialogue simulator, with significant cases observed.
\end{itemize}


\section{Problem Formulation}
\label{sec:problem_formulation}

Naturally, an {\MultiParty} sample consists of multiple roles and utterances. We assume a scene description can be constructed for an arbitrary {\MultiParty} sample, which contains information on participating roles, the conversation topic, location or other contexts. Utterances appear in an interleaved manner and belong to different roles. For simplicity, we assume the adjacent utterances can not belong to the same role.

As the prerequisite of methodology derivation, here we first propose some variable definitions, to formulate the {\MultiParty} problem. Given a {\MultiParty} sample, there are maximally $L$ roles and $T$ utterances; we further assume $s$ denotes the scene description, $u_t$ denotes the content of the $t$-th utterance, while $r_t$ denotes the role index that the $i$-th utterance belongs to:
\begin{equation*}
r_t = r(u_t) \in [0, \cdots, L-1], t \in [0, \cdots, T-1]
\end{equation*}
For abbreviation, we use the following shortcut variable to indicate the utterance sequence:
\begin{equation}
\label{eq:utterance_seq}
    \{u\}_{0:t} := \{ u_t, t \in [0, \cdots, t] \}
\end{equation}

\begin{figure*}[!htbp]
  \includegraphics[width=1\linewidth]{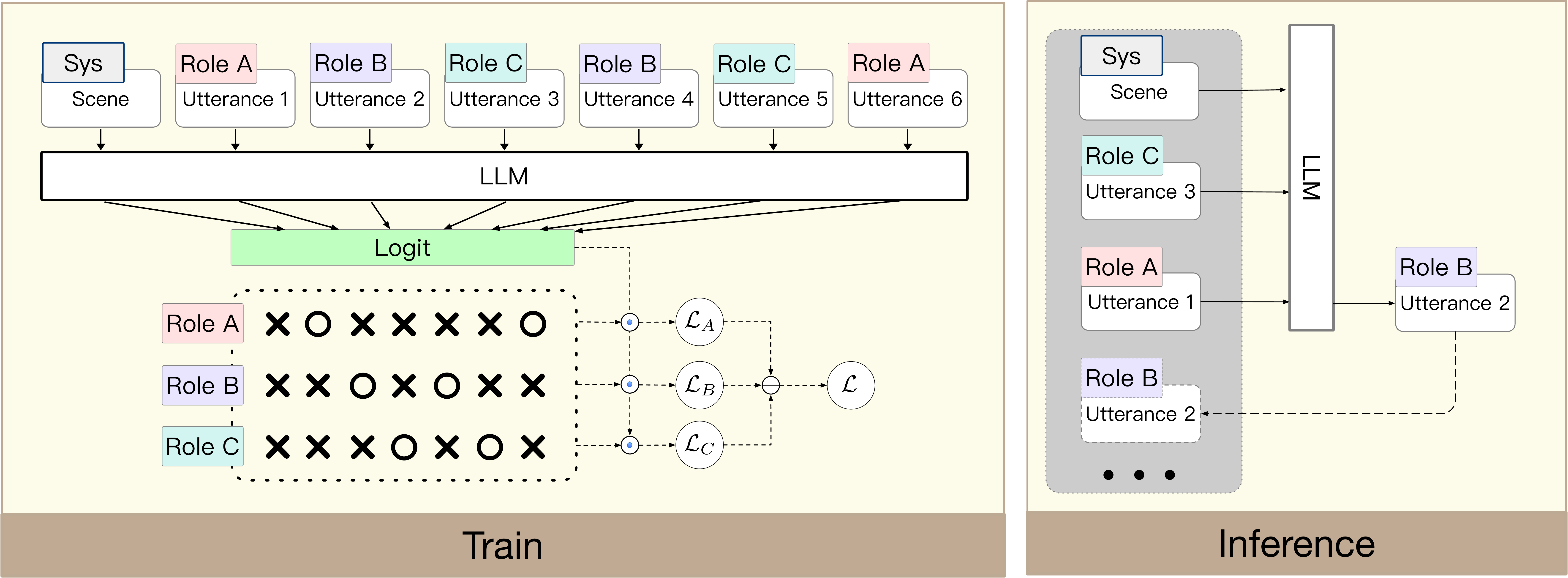} 
  \caption {The entire framework of {\ModelName}. \\ 
  Train: LLM performs a forward pass to obtain the logits of the multi-party dialogue. For each role, the fine-tuning loss is calculated with inactive role parts masked. \\ 
  Inference: LLM generates the next utterance given the system prompt and historical utterances. New generated utterances can be appended to the end of dialogue and inference can continue further.}
  \label{fig:framework}
\end{figure*}

\section{Method}
\label{sec:method}

In this section, we propose a straightforward but effective LLM-based approach to solve the {\MultiParty} problem. We demonstrate training and inference details, then provide further strategies to convert the model to a {\MultiParty} simulator. Figure \ref{fig:framework} indicates our methodology. 

\subsection{Training}
\label{sec:general_train}



Figure \ref{fig:framework} (Left) visualizes the training methodology. Similar to the conventional LLM training, logit of the {\MultiParty} textual input is obtained by a forward pass of LLM. For each role of the sample, we calculate its Supervised Fine-Tuning (SFT) loss by masking out the tokens' corresponding utterances of the system and all other roles \footnote{If the active role has utterance1, it might be better to also mask this part; however, here we just omit this detail for demonstration clarity.}. We average each role's loss to obtain the entire training loss:
 
\begin{equation}
    \mathcal{L} = - \frac{1}{L} \sum_{i=1}^{L} \log \left[ \text{P}(\{u\}^{r = i}_{0:T} \mid s, \{u\}^{r \neq i}_{0:T}) \right]
    \label{eq:sft}
\end{equation}
in which $\log \left[ \text{P}(\cdot) \right]$ indicates the log probability calculated by the current LLM, $\{u\}^{r = i}$ and $\{u\}^{r \neq i}$ are abbreviations of the utterance sequence whether belongs to and not to the $i$-th role: 
\begin{align}
    \{u\}^{r = i}_{0:T} := \{ u_t, t \in [0, \cdots, T] \mid r(u_t) = i \} \\
    \{u\}^{r \neq i}_{0:T} := \{ u_t, t \in [0, \cdots, T] \mid r(u_t) \neq i \}
\end{align}


\subsection{Inference}
\label{sec:general_inference}

During the inference stage, {\ModelName} is first assigned with the current role, then generates its utterance grounded by the system prompt and previous utterances:
\begin{align}
u_{t} \leftarrow \text{LLM}(s, \{u\}_{0:t-1}, r_t) \label{eq:general_inference}
\end{align}
where the left arrow means LLM generation. We then append $u_{t}$ into the end of dialogue and proceed incrementally (if needed). The inference pipeline is shown in Figure \ref{fig:framework} (Right). 



\subsection{The {\MultiParty} Simulator}
\label{sec:simulator}
A more interesting and intriguing application might be the {\MultiParty} simulation, where a series of speaking roles and their utterances are needed to generate sequentially, with some pretended scene description and utterances. Such a simulator can be applied in debate rehearsal, show script auto-writing, or meta-universe creation. Note this situation is different and more complicated than the inference stage introduced in Subsection \ref{sec:general_inference}, where the speaking role is foreknown. To build a {\MultiParty} simulator, the next-speaker prediction or recognition is also needed; it is also important that the model can adapt with some specific role description and portrays different characteristics or personas.

We integrate the above tasks into a comprehensive task and find that the LLM fine-tuning framework can handle it efficiently, with only minor methodological revisions. Motivated by the difference between centralization and decentralization architectures, we propose the \textit{Speaker Predictor} and \textit{Silence Switcher} strategies respectively, which are demonstrated in the following subsections.



\subsubsection{Speaker Predictor} 

We re-paraphrase the next speaker role ($r_t$) as part of generation during inference, and correspondingly unmask its loss during training. In such a manner, the LLM is trained to generate first $r_t$ and then $u_t$. 
\begin{align}
r_t, u_{t} \leftarrow \text{LLM}(s, \{u\}_{0:t-1}) \label{eq:speaker_predict}
\end{align}
By such fine-tuning, only a single LLM object is needed to simulate the {\MultiParty}, which is in charge of generating both roles and utterances of different turns, in an unidirectional and causal manner.

\subsubsection{Silence Switcher}

In this strategy, the LLM is still grounded with the current role but also allowed to possibly generate `<s>', a special token representing the `silence'. The simulator then becomes a multi-agent framework where different LLMs (or one LLM with dynamically switching role prompts) portray different roles. 

Upon each utterance generation, we allow each LLM to speculate its possibility of `silence', and choose the one with the minimum likelihood as the current speaker:
\begin{align}
r_{t} = \arg\min_{i} \log \left[ \text{P} (\text{<s>} \mid s, \{u\}_{0:t-1}, r_t=i) \right] \label{eq:silence_min_logprob}
\end{align}
Then the LLM is called again to generate the utterance content $u_{t}$ based on Equation \ref{eq:general_inference}, and the dialogue continues incrementally until the maximum turn number is reached. 


\section{Experiments}
\label{sec:experiment}

In this section, we first provide the experimental settings, then the training results including dialogue generation and speaker prediction, and finally some thorough discussions.

\begin{table*}[htbp!]
\caption{Details of Training Datasets.}
\label{tab:datasets}

\begin{center}
\small
\renewcommand{\arraystretch}{0.9}
\resizebox{\textwidth}{!}{
\begin{tabular}{c|lcccc}
\toprule 
Split & Dataset Name & Task & \# Clips & \# Utterance & \# Utterance per Clip \\ 
\toprule
\multirow{7}{*}{Train} 
 & Friends \cite{yangEndEndPersonalizedHumorous2019} & Show Scripts & 5324 & 63724 & 11.97 \\
 & Chat-Haruhi \cite{liChatHaruhiRevivingAnime2023} & Show Scripts & 184561  & 1826920 & 9.90 \\
 & Chat-Suzumiya \cite{liChatHaruhiRevivingAnime2023} & Show Scripts & 122768  & 1210002 & 9.86 \\
& Tv dialogue$^{\ast}$ & Show Scripts & 139797 & 1400704 & 10.02 \\
 & British Parliamentary \cite{liangDebatrixMultidimensionalDebate2024} & Debate & 43 & 463 & 10.77 \\              
& IQ2US \cite{zhangConversationalFlowOxfordstyle2016} & Debate  & 2660 & 26562 & 9.99 \\

& Annotated US Supreme Court Arguments$^{\blacktriangle}$ & Debate & 4739  & 47312 & 9.98 \\
\midrule 
\multirow{2}{*}{Test} 
 & Friends \cite{yangEndEndPersonalizedHumorous2019} & Show Scripts & 592 & 7086 & 11.97 \\
 & Game of Thrones$^{\blacktriangledown}$ & Show Scripts & 2086 & 21237 & 10.18 \\
\bottomrule 
\end{tabular}
}
\end{center}

${\ast}$: \url{https://huggingface.co/datasets/sedthh/tv_dialogue} \\
${\blacktriangle}$: \url{https://www.kaggle.com/datasets/jameslabadorf/us-supreme-court-arguments-20172021} \\
${\blacktriangledown}$: \url{https://www.kaggle.com/datasets/albenft/game-of-thrones-script-all-seasons}

\end{table*}

\subsection{Settings}
\label{sec:exp_setting}

\subsubsection{Datasets Details}
\label{sec:dataset_detail}

We collect substantial {\MultiParty} datasets most of which belong to two main categories: the show scripts \cite{yangEndEndPersonalizedHumorous2019,liChatHaruhiRevivingAnime2023} and debates records \cite{liangDebatrixMultidimensionalDebate2024,zhangConversationalFlowOxfordstyle2016}. Among these, we divide the `Friends' dataset into the training and test sets with the same split fraction as \cite{yangEndEndPersonalizedHumorous2019}, such that some of their experimental results can be directly compared. We also use the entire `Game of Thrones' dataset \footnote{\url{https://www.kaggle.com/datasets/albenft/game-of-thrones-script-all-seasons}} as the test set, to test the zero-shot ability. We summarize the statistics and configuration of training datasets in Table \ref{tab:datasets}. We limit each sample contains mostly 10 utterances and divide the clip into multiple parts which is longer than that.

We further illustrate the experimental details to test different aspects of model capabilities:
\begin{itemize}
    \item Test: We select the scene description and the first utterance of each sample of the Friends test test, and let the model extend the {\MultiParty} by generating more utterances. 
    \item Generalization: We manually write the scene description and the first utterance according to the Friends scenario; since the model already learns the roles' characteristics and talking corpus through the training dataset, this approach tests the model completion ability given arbitrary scene and previous utterances.
    \item Zero-Shot: we select the beginning utterances (maybe 2$\sim$3) of the Game of Thrones (GOT) samples (not covered by the training set) and manually write descriptive scenes. This approach tests the model's zero-shot ability given unseen role definitions and utterances.
\end{itemize}


\begin{table*}[t!]
\caption{Response Quality Evaluation on the Test Set of Friends$^{\ast}$. Values with bold indicate the best results while values with underline indicate the second-best results.}
\label{tab:response_quaility}

\centering
\resizebox{\textwidth}{!}{%
\begin{tabular}{c|c|c|c|c|cccc}
    \toprule
    \multicolumn{3}{c|}{\multirow{2}[4]{*}{\textbf{Method}}} & \multirow{2}[4]{*}{\textbf{Size}} & \multicolumn{1}{c|}{\textbf{GPT-4}} & \multicolumn{4}{c}{\textbf{Human Annotation}} \\
\cmidrule{5-9}    \multicolumn{3}{c|}{} &       & \multicolumn{1}{c|}{Score $\uparrow$} & Fluency $\uparrow$ & Consistency $\uparrow$ & Entertainment $\uparrow$ & Total $\uparrow$ \\
    \midrule
    \multicolumn{2}{c|}{\multirow{5}[2]{*}{Zero-shot}} & Prompt+Llama3 & 70B   & 7.89 ± 1.11 & 7.8 ± 1.0 & 7.3 ± 0.9 & 7.7 ± 0.9 & 22.8 ± 1.9 \\
    \multicolumn{2}{c|}{} & Prompt+Qwen2 & 72B   & 7.69 ± 1.43 & 6.7 ± 1.6 & 6.9 ± 0.7 & 6.7 ± 0.7 & 20.3 ± 2.3 \\
    \multicolumn{2}{c|}{} & Prompt+Mistral & 8x7B   & 7.61 ± 1.55 & 6.9 ± 0.9 & 7.3 ± 1.0 & 7.1 ± 1.1 & 21.3 ± 2.5 \\
    \multicolumn{2}{c|}{} & Prompt+Deepseek-v2 & 236B  & 7.78 ± 1.39 & 6.6 ± 1.2 & 6.8 ± 1.1 & 6.1 ± 0.6 & 19.5 ± 2.7 \\
    \multicolumn{2}{c|}{} & Prompt+GPT-4 & N/A   &\textbf{8.32} ± 1.26  & 7.9 ± 0.6 & 7.7 ± 0.8 & 6.9 ± 0.6 & 22.5 ± 1.9 \\
    \midrule
    \multirow{6}[4]{*}{\begin{sideways}Fing-Tuning\end{sideways}} & \multirow{3}[2]{*}{\begin{sideways}\footnotesize{Llama3}\end{sideways}} & VanillaSFT & \multirow{3}[2]{*}{8B} &  7.01 ± 2.29      & 7.6 ± 0.9     & 7.0 ± 1.1     & 7.8 ± 1.0     & 22.4 ± 1.9 \\
          &       & MuPaS-Speaker(ours) &       & 8.02 ± 2.14       & \underline{8.2} ± 0.7     & 7.9 ± 0.7     & \underline{8.1} ± 0.8     & {24.2} ± 1.7 \\
          &       & MuPaS-Silence(ours) &       & \underline{8.07} ± 1.84      & \underline{8.2} ± 0.6     & \underline{8.0} ± 0.9     & \textbf{8.3} ± 0.8     & \textbf{24.5} ± 1.6 \\
\cmidrule{2-9}          & \multirow{3}[2]{*}{\begin{sideways}\footnotesize{Qwen2}\end{sideways}} & VanillaSFT & \multirow{3}[2]{*}{7B} & 7.22 ± 1.61 & 7.6 ± 0.9 & 7.3 ± 0.8 & 7.5 ± 0.7 & 22.4 ± 2.0 \\
          &       & MuPaS-Speaker(ours) &       & 7.78 ± 1.38 & \textbf{8.3} ± 0.8 & \textbf{8.1} ± 0.7 & \underline{8.1} ± 0.7 & \textbf{24.5} ± 2.0 \\
          &       & MuPaS-Silence(ours) &       & 7.34 ± 1.49 & 8.1 ± 0.7 & \underline{8.0} ± 0.6 & {7.9} ± 0.8 & {24.0} ± 1.8 \\
    \bottomrule
    \end{tabular}%
}

${\ast}$: We do not include results of non-LLM methods since most of these models are not open-sourced.
\end{table*}

\subsubsection{Baselines}

\begin{itemize}
\item Previous non-LLM based works on {\MultiParty}, such as MIDS \cite{yangEndEndPersonalizedHumorous2019}, SI-RNN \cite{zhangAddresseeResponseSelection2017} and Static/Dynamic-ADR \cite{ouchiAddresseeResponseSelection2016}.
\item The prompt-based approach. We achieve so by converting the {\MultiParty} problem into a single-turn instruction following task, in which we concatenate historical utterances into a single user query, and write an extra instruction to let LLM generate {\MultiParty} response grounded by multi-party history.

\item The vanilla SFT method ({\vanillaFT}) on LLM which also concatenates historical utterances as the query, and labels the ground-truth utterance as the target text.
\end{itemize}

For LLM-based baselines, we examine Qwen2-7B-Instruct \cite{qwen2techreport2023}, Llama3-8B-Instruct \cite{llama3modelcard}, Deepseek-v2 \cite{deepseek2024deepseekv2technicalreport} and GPT-4 \cite{openai2024gpt4technicalreport}. Qwen2-7B-Instruct and Llama3-8B-Instruct are also set as our base model. We experiment with {\ModelName}-Speaker and {\ModelName}-Switcher corresponding to \textit{the Speaker Predictor} and \textit{the Silence Switcher} strategies proposed in Section \ref{sec:simulator}.

\subsubsection{Hyper-parameters}

The learning rate is $5.0e-6$, the training batch size is 32 and the sequence window length is 2048. The training epoch is set to 2. We perform the training experiment in LlamaFactory \cite{zheng2024llamafactory}, running by 8 A100 GPUs. We use the AdamW optimizer with the cosine scheduler of learning rate and decay of 0.01. We first train the model with some open-domain dialogue and reasoning datasets then conduct some detailed downstream fine-tuning tasks.

\begin{table*}[h!]
    \caption{Simulated examples of {\ModelName} on the test test of Friends. Text in brown is the pre-determined input into our model, and the other texts are generated by {\ModelName}.} 
    \label{tab:friends_case}
    
    \centering
    \resizebox{\textwidth}{!}{%
    \begin{tabular}{c|l}
        \Xhline{2\arrayrulewidth}
        \multicolumn{2}{c}{\textbf{Test}}  \\ 
        \hline
        \textcolor{brown}{Scene} & \makecell[l]{\textcolor{brown}{Central Perk, Ross, Phoebe, and Chandler are there.}}  \\
\hline
       \textcolor{brown}{Ross} & \makecell[l]{\textcolor{brown}{Oh, wow! I should get going. I-I got a date tonight.}}   \\
\hline
      Phoebe & \makecell[l]{Ohhh, nice! where're you going?}    \\ 
\hline
      Ross & \makecell[l]{Uh, the American Museum of Natural History.}   \\
      \hline
      Chandler & \makecell[l]{Oh yeah, that place where people go with their dates to look at fossils, not each other.}  \\
      \hline
      Phoebe & \makecell[l]{So what's she like?}  \\
      \hline
      Ross & \makecell[l]{Uh, she has a masters degree and she's into dinosaurs. So I mean, she's kind of my soulmate.}  \\
        \Xhline{2\arrayrulewidth}
        \multicolumn{2}{c}{\textbf{Generalization}}  \\ 
        \hline
        \textcolor{brown}{Scene} & \makecell[l]{\textcolor{brown}{Rachel, Joey, and Monica are at the bank.}}  \\
\hline
     \textcolor{brown}{Rachel} & \makecell[l]{\textcolor{brown}{Darn it,why is the line so long?}}   \\
\hline
     Joey & \makecell[l]{I know! why do they always put all the slow people at the front of the line?}   \\
     \hline
     Monica & \makecell[l]{Thats not fair.}   \\
     \hline
     Rachel & \makecell[l]{Yeah, its true. Look at that woman, shes been here forever.}   \\
        \Xhline{2\arrayrulewidth}
    \end{tabular}
    }

\end{table*}

\begin{figure}[t]
  \includegraphics[width=1.0\linewidth]{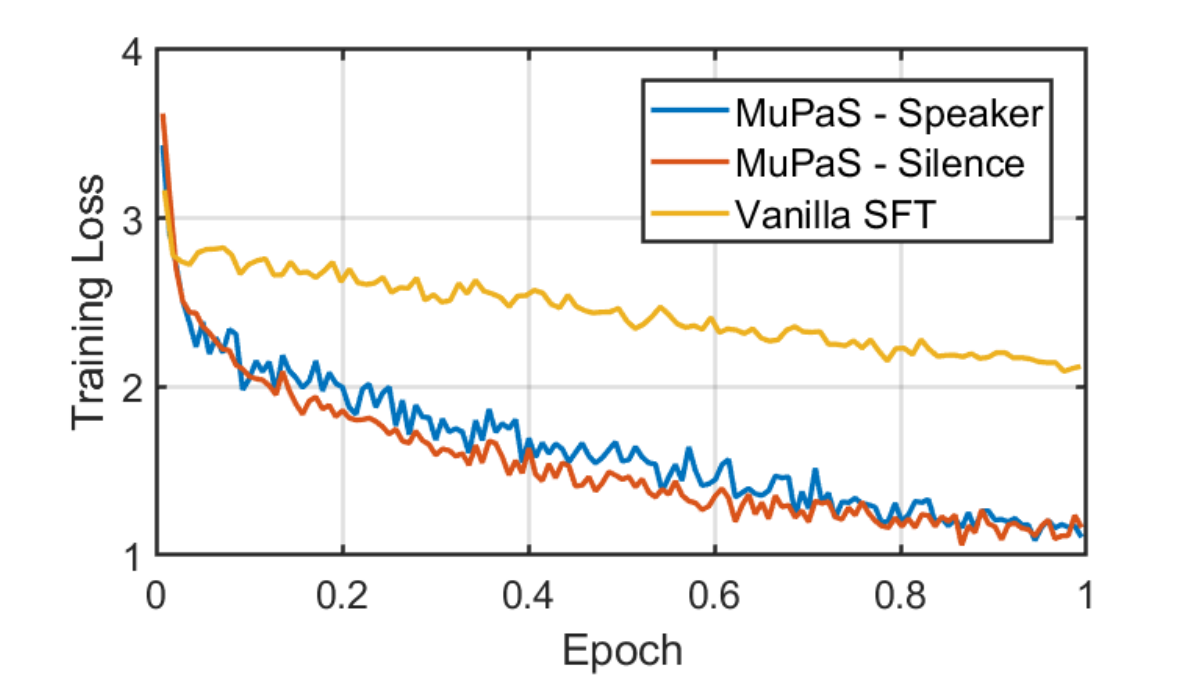} 
  \caption {Training loss curves of {\ModelName}-Speaker, {\ModelName}-Silence and {\vanillaFT}. To make the transient dynamics clear, only the first epoch (totally 2) is exhibited.}
  \label{fig:losses}
\end{figure}

\subsection{Results}

Figure \ref{fig:losses} presents the loss curves for the Speaker Predictor and Silence Switcher methods in {\ModelName}. Initially, both approaches exhibit high loss as the instruction-based LLM transitions from a two-party to a multi-party paradigm. However, the loss decreases rapidly, converging to a stable value by the end of the training period, indicating that the LLM can effectively learn to engage in {\MultiParty} dialogues when provided with sufficient data. Furthermore, the training labels used in Silence Switcher are more aligned with traditional SFT, leading to a faster decrease in loss during the initial stages of training and more stable performance curves. In contrast, the training approach of Vanilla SFT results in the loss of some original conversational information, causing slower progress and higher loss even after one epoch.

\subsubsection{Quantitative Results}

Table \ref{tab:response_quaility} lists the quality assessment results of {\MultiParty} responses, by the manners of LLM auto-evaluation and human annotation. In this evaluation, the automatic assessment of the model utilizes the advanced GPT-4, which assigns scores ranging from 0 to 10 based on the fluency of the dialogue and adherence to the character traits established in Friends. Additionally, the manual assessment is conducted by trained professionals and individuals with prior experience watching Friends. They evaluate the dialogue on three criteria: Fluency, Consistency, and Entertainment, with each criterion having a maximum score of 10. 

Our model outperforms other baselines on both assessment approaches. While the automatic evaluation by GPT-4 considers its own dialogue generation quality to be the best, our method still achieves the second-highest score. In contrast, human evaluations show that our approach outperforms all baselines across various metrics. Although this work only tests {\ModelName} on a 7B or 8B base model, its {\MultiParty} performance is even better than baselines with a much stronger base model, \textit{e.g.} 70B or 236B open-sourced instruct models, or the top closed-sourced LLM such as GPT-4.


\begin{table*}[h!]
    \caption{Two typical zero-shot simulated examples of {\ModelName} on Game of the Throne (GOT). Text in brown is the pre-determined input into our model, and the other texts are generated by model.} 
    \label{tab:got_case}
    \centering
    \small
    \resizebox{\textwidth}{!}{%
    \begin{tabular}{c|l}
\Xhline{2\arrayrulewidth} 
\textcolor{brown}{Scene} & \textcolor{brown}{\makecell[l]{\textbf{Jon Snow}, raised as the illegitimate son of Ned Stark, is honorable, brave, and known for his sense of duty. He recently discovered \\ he is the legitimate son of Rhaegar Targaryen and Lyanna Stark, making him Aegon Targaryen—the true heir to the Iron Throne.\\
\textbf{Daenerys Targaryen}, known as the Mother of Dragons, is determined, strong-willed, and believes she is the last Targaryen heir.\\ She has been on a quest to reclaim the Iron Throne, which she sees as her birthright.\\
They are in a quiet room at Winterfell after the Battle of Winterfell. The atmosphere is tense but intimate, with the soft glow of \\ candlelight. They are sitting near a fireplace, and the mood is heavy with unspoken words.}}  \\
\toprule
\textcolor{brown}{Jon Snow} & \textcolor{brown}{\makecell[l]{Daenerys, there's something important I need to tell you.}}  \\
\hline
\textcolor{brown}{Daenerys} & \textcolor{brown}{\makecell[l]{What is it? You look troubled.}}   \\
\hline
Jon Snow & \makecell[l]{I've discovered my true identity. I am the son of Rhaegar Targaryen and Lyanna Stark. I am Aegon Targaryen, the true heir to \\ the Iron Throne.}   \\
\hline
Daenerys  & \makecell[l]{(whispering) You can't be serious.}   \\
\hline
Jon Snow & \makecell[l]{I am. My mother's ring was found on my nightstand. It has the Targaryen dragon symbol engraved on it.}   \\
\hline
Daenerys  & \makecell[l]{(voice shaking) That means... I'm not the only Targaryen left.}   \\
\hline
Jon Snow & \makecell[l]{Yes, but that doesn't change the fact that you're the rightful heir.}   \\
\hline
Daenerys  & \makecell[l]{And you're saying all this because...?}   \\
\hline
Jon Snow & \makecell[l]{Because I want you to rule the Seven Kingdoms.}   \\
\hline
Daenerys  & \makecell[l]{(voice breaking) And why would you want that for me?}   \\
\hline
Jon Snow & \makecell[l]{Because I believe in you, Daenerys. You have the heart, the courage, and the determination to lead the realm. You have \\ the dragons, and now you have the North. Together, we can build a brighter future.}   \\
\bottomrule
\toprule
\textcolor{brown}{Scene} & \textcolor{brown}{\makecell[l]{In King's Landing, the capital of the Seven Kingdoms of Westeros, Tywin Lannister has convened an urgent meeting of the Small\\ Council to address the kingdom's escalating financial deficits and the rise of rebellions in various regions. Characters:\\
\textbf{Tywin Lannister}: The patriarch of House Lannister and Hand of the King, Tywin is a formidable statesman known for his ruthless\\ efficiency and unwavering commitment to his family's legacy. He is stern, calculating, and intolerant of failure. Tywin's primary\\ motivation is to preserve the power and influence of House Lannister, ensuring its dominance over the Seven Kingdoms.\\
\textbf{Tyrion Lannister}: Nicknamed "The Imp", Tywin's youngest son. Despite being marginalized due to his dwarfism, he possesses a\\ sharp intellect, wit, and a keen understanding of politics and human nature. Tyrion seeks recognition and respect, often advocating\\ for pragmatic solutions. His motivations include proving his worth to his father and promoting more just and effective governance.\\
\textbf{Varys}: Known as "The Spider", the Master of Whisperers. A eunuch originally from Essos, he has built an extensive network of\\ informants across the realm. Varys is enigmatic, patient, and perceptive, valuing the stability of the realm above personal ambition.\\ His primary motivation is to maintain peace and protect the common people from the chaos of war and misrule.\\
\textbf{Petyr Baelish}: Referred to as "Littlefinger", the Master of Coin. Born to a modest noble house, he is cunning, ambitious, and adept\\ at manipulation. Petyr is a master of political intrigue, using his financial acumen and strategic alliances to climb the social ladder.\\ His motivations are self-serving; he thrives on chaos, seeing it as an opportunity to amass more power and wealth.}} \\
        \toprule
        \textcolor{brown}{Tywin Lannister} & \textcolor{brown}{We face a dire situation. The kingdom is bleeding gold, and the recent uprisings are a direct threat to our power.} \\
        \hline
        Tyrion Lannister & Perhaps we should consider reforming the tax system. A fairer system could increase revenue and appease the common folk. \\
        \hline
        Varys & Tax reform may help, but we must also address the grievances of the people. They are more likely to rebel if they feel unheard. \\
        \hline
        Petyr Baelish & I agree with Varys. We need to create an illusion of stability and prosperity. We can invest in some public works to distract them. \\
        \hline
        Tywin Lannister & And what about our foreign affairs? We cannot afford to neglect our alliances or provoke new enemies. \\
        \hline
        Varys & A wise reminder, Lord Hand. We must tread carefully in matters of diplomacy and defense. \\
        \hline
        Tywin Lannister & \makecell[l]{In addition to these concerns, I propose a restructuring of the royal guard. We need a more efficient and loyal protection force to\\ ensure the safety of the realm.} \\
        \hline
        Tyrion Lannister & I believe that focusing on these internal issues is crucial, but we must not ignore the potential threats from beyond our borders. \\
        \hline
        Petyr Baelish & Exactly, Lord Hand. We must be vigilant against any external pressures that could destabilize our rule. \\
        \hline
        Tywin Lannister & \makecell[l]{I appreciate your input, my lords. Let us proceed with these proposals and make necessary adjustments to restore the\\ prosperity and unity of the Seven Kingdoms.} \\
        \hline
        Varys & Yes, my Lord Hand. We shall work together to overcome these challenges. \\
        \hline
        Tyrion Lannister & And so, we embark on a new chapter of governance, guided by reason and a commitment to the greater good. \\
        \hline
        Varys & The realm owes you a debt of gratitude, my Lord Hand. \\
        \hline
        Tyrion Lannister & Thank you, my Lords. Together, we shall shape the destiny of Westeros. \\
\bottomrule
    \end{tabular}
    }
\end{table*}

\subsubsection{Results of Multi-Party Simulation}

We employ the trained model to build a {\MultiParty} simulator. We examine the performance of {\ModelName} across different configurations proposed in Subsection \ref{sec:exp_setting}, and observe reasonable performance. Table \ref{tab:friends_case} exhibits both typical test-set and generalization cases. {\ModelName} can produce fluent, consistent and high-quality utterances, no matter whether the scene description and previous utterances are sampled from test set directly or written manually, indicating its high robustness. Furthermore, it can be observed that each utterance's style matches the role's characteristics. For example, Ross is more prone to talk about academic topics while Monica cares about fairness. 


We observe astonishing results for zero-shot experiments, in which the entire set of story contexts and role styles have not been studied by the model fine-tuning\footnote{Yet it might nevertheless have partially related knowledge from the pre-training phase.}. In this case, we manually input more content of scene description which includes the role introductions, their location, topics and other contexts. {\ModelName} generates fluent and interesting dialogues between multiple roles even if it does not know well about them before prompted. See Table \ref{tab:got_case} for two cases `Jon Snow is preparing to tell Daenerys his true identity' and `Tywin Lannister, Tyrion Lannister, Varys and Petyr Baelish are having a council meeting'. 




\begin{table*}[h!]
  \caption{Ablation Studies. Loss is averaged from the original step-wise values of the second epoch, after the loss curve becomes stable. Accuracy is the abbreviation of the next-speaker prediction accuracy, which is the same term as reported in Table \ref{tab:speaker_prediction}.}
  \label{tab:ablation}%
  
  \centering
  \small
  \renewcommand{\arraystretch}{1.0}{
  \begin{tabular}{c|cl|cc|cccc}
    \toprule
    \multicolumn{3}{c|}{\multirow{2}[4]{*}{\textbf{Method}}} & \multicolumn{2}{c|}{\textbf{Auto-Metrics}} & \multicolumn{4}{c}{\textbf{Human Annotation}} \\
    \cmidrule{4-9}    
    \multicolumn{3}{c|}{} & Loss $\downarrow$  & Accuracy $\uparrow$  & Fluency $\uparrow$ & Consistency $\uparrow$ & Entertainment $\uparrow$ & Total $\uparrow$ \\
    \midrule
    \multirow{8}[4]{*}{\begin{sideways}Speaker Predictor\end{sideways}} 
    & \multirow{4}[2]{*}{\rotatebox{90}{\footnotesize Llama3}} 
    & utterance-level loss & 1.36     & 77.15     & 7.5 ± 1.2     & \textbf{7.9} ± 0.9     & 8.0 ± 1.0     & 23.4 ± 1.6 \\
    & & without scene & 1.15     & 77.49     & 7.8 ± 0.9     & 7.4 ± 1.1     & 7.6 ± 1.0     & 22.8 ± 1.8 \\
    & & from base & 3.55     & 55.07     & 5.9 ± 1.6     & 5.6 ± 1.3     & 6.5 ± 1.3     & 18.0 ± 2.8 \\
    & & \textbf {\ModelName} & \textbf{0.92}     & \textbf{81.38}     & \textbf{8.2} ± 0.7     & \textbf{7.9} ± 0.7     & \textbf{8.1} ± 0.8     & \textbf{24.2} ± 1.7 \\
    \cmidrule{2-9}          
    & \multirow{4}[2]{*}{\rotatebox{90}{\footnotesize Qwen2}} 
    & utterance-level loss & 1.42 & 69.42 & 7.4 ± 1.0 & 7.8 ± 0.9 & 7.5 ± 1.1 & 22.7 ± 2.0 \\
    & & without scene & 1.54 & 72.13 & 7.6 ± 1.1 & 7.4 ± 1.2 & 7.8 ± 0.9 & 22.8 ± 2.1 \\
    & & from base & 1.34 & 77.53 & 7.6 ± 1.0 & 7.9 ± 1.0 & \textbf{8.1} ± 0.6 & 23.6 ± 2.1 \\
    & & \textbf {\ModelName} & \textbf{1.11} & \textbf{81.76} & \textbf{8.3} ± 0.8 & \textbf{8.1} ± 0.7 & \textbf{8.1} ± 0.7 & \textbf{24.5} ± 1.8 \\
    \midrule
    \multirow{8}[4]{*}{\begin{sideways}Silence Switcher\end{sideways}} 
    & \multirow{4}[2]{*}{\rotatebox{90}{\footnotesize Llama3}} 
    & utterance-level loss & 1.63     & 76.99     & 7.8 ± 0.9     & 7.5 ± 0.9     & 7.8 ± 1.0     & 23.1 ± 1.6 \\
    & & without scene & 1.20     & 78.34     & 8.0 ± 0.8    & 7.3 ± 1.2     & 7.7 ± 0.9     & 23.0 ± 1.9 \\
    & & from base & 3.76     & 57.60     & 5.8 ± 1.2     & 6.1 ± 1.1     & 6.6 ± 1.2     & 18.6 ± 2.6 \\
    & & \textbf {\ModelName} & \textbf{1.00}     & \textbf{81.21}     & \textbf{8.2} ± 0.6     & \textbf{8.0} ± 0.9     & \textbf{8.3} ± 0.8     & \textbf{24.5} ± 1.6 \\
    \cmidrule{2-9}          
    & \multirow{4}[2]{*}{\rotatebox{90}{\footnotesize Qwen2}} 
    & utterance-level loss & 1.76 & 59.46 & 6.5 ± 1.3 & 6.7 ± 1.2 & 7.0 ± 0.8 & 20.2 ± 2.5 \\
    & & without scene & 1.82 & 58.61 & 7.1 ± 1.2 & 5.5 ± 1.6 & 6.6 ± 1.3 & 19.2 ± 2.3 \\
    & & from base & 1.20 & 76.86 & 7.8 ± 0.6 & 7.8 ± 0.9 & \textbf{8.1} ± 0.9 & 23.7 ± 1.4 \\
    & & \textbf {\ModelName} & \textbf{1.12} & \textbf{80.07} & \textbf{8.1} ± 0.7 & \textbf{8.0} ± 0.6 & 7.9 ± 0.8 & \textbf{24.0} ± 1.8 \\
    \bottomrule
  \end{tabular}%
  }

\end{table*}

\subsubsection{Speaker Prediction Accuracy}
Table \ref{tab:speaker_prediction} shows the speaker prediction accuracy on the Friends test set. It can be observed that methods relying on LLM prompting generally achieve relatively low accuracy, ranging from 61.49\% for Deepseek-v2 to 72.47\% for GPT-4. After applying Vanilla Supervised Fine-Tuning, there is a noticeable improvement in accuracy. In addition, traditional approaches that rely on multi-party dialogue modeling tend to perform better in this task surprisingly, as they are specifically designed and trained to handle the final round of dialogue. Nevertheless, our {\ModelName} method, without making any special adjustments for the final round, consistently achieves an accuracy over 80\%, outperforming all previous studies.

\begin{table}[h]
\caption{Results of the next-speaker prediction on the test set of Friends. The maximum number of roles is 7.}
\label{tab:speaker_prediction}

\centering
\renewcommand{\arraystretch}{1.0}
\setlength{\tabcolsep}{3pt}  

\begin{tabular}{>{\centering\arraybackslash}m{0.8cm} c c c c}
\toprule
\textbf{} & \textbf{Method} & \textbf{Base} & \textbf{Size} & \textbf{Accuracy (\%)} \\ 
\midrule
\multirow{5}{*}{\rotatebox[origin=c]{90}{\footnotesize Non-LLM}} 
 & Static-ADR$^{\ast}$ \cite{ouchiAddresseeResponseSelection2016} & - & - & 74.37 \\
 & Dynamic-ADR$^{\ast}$ \cite{ouchiAddresseeResponseSelection2016} & - & - & 76.48 \\
 & SI-RNN$^{\ast}$ \cite{zhangAddresseeResponseSelection2017} & - & - & 76.50 \\
 & MIDS (no context)$^{\ast}$ \cite{yangEndEndPersonalizedHumorous2019} & - & - & 69.94 \\
 & MIDS$^{\ast}$ \cite{yangEndEndPersonalizedHumorous2019} & - & - & 79.32 \\
\midrule
\multirow{5}{*}{\rotatebox[origin=c]{90}{\footnotesize Zero-Shot}} 
 & Prompt  & Mistral & 8x7B & 62.84 \\
 & Prompt  & Deepseek-v2 & 236B & 61.49 \\
 & Prompt & Llama3 & 70B & 65.37 \\
 & Prompt & Qwen2 & 72B & 67.74 \\
 & Prompt & GPT-4 & N/A & 72.47 \\
\midrule
\multirow{6}{*}{\rotatebox[origin=c]{90}{\footnotesize Fine-Tuning}} 
 & {\vanillaFT} & Llama3 & 8B & 74.66 \\
 & {\vanillaFT} & Qwen2 & 7B & 75.00 \\
  & {\ModelName} - Speaker (ours) &  Llama3 & 8B & \textbf{81.38} \\
 & {\ModelName} - Silence (ours) & Llama3 & 8B & \textbf{80.21} \\
 & {\ModelName} - Speaker (ours) & Qwen2 & 7B & \textbf{81.76} \\
 & {\ModelName} - Silence (ours) & Qwen2 & 7B & \textbf{80.07} \\
\bottomrule
\end{tabular}

$\ast$: we directly obtain the results from the original paper \cite{yangEndEndPersonalizedHumorous2019}.
\end{table}

\begin{table}[t]
\caption{Benchmarks of generalized capabilities.}
\label{tab:addlabel}
\small
  \centering
    \begin{tabular}{c|cc}
    \toprule

Metrics & Llama3-8B-Instruct & MuPaS \\

    \midrule
    MMLU  & \textbf{67.51} & 66.23 \\
    BBH   & \textbf{40.65} & 33.77 \\
    TriviaQA & \textbf{61.83} & 61.47 \\
    GSM8K & 35.1  & \textbf{43.14} \\
    TruthfulQA & 37.45 & \textbf{44.33} \\
    \bottomrule
    \end{tabular}
\end{table}

\subsection{Ablation Study}

To investigate the impact of different model components on overall performance, this section explores the effects of modifying the conditions for the speaker and silence models. The following approaches are employed:

\begin{itemize}
\item Utterance-level loss: For each data instance, only one speaker's utterances are randomly selected for training, allowing for an analysis of how different {\MultiParty} learning strategies affect the training process.
\item Without scene: The system prompt descriptions of roles and context are removed, with the model trained solely on dialogues between speakers.
\item From base: The model is trained directly from the Llama3-8B or Qwen2-7B  base model, rather than from an SFT-finetuned Instruct model.
\end{itemize}

The evaluation results, as shown in Table \ref{tab:ablation} , indicate that altering any training condition leads to an increase in loss at the end of the first epoch, signaling a slowdown in training. Furthermore, both the accuracy of role prediction and the quality of content generation deteriorate. A cross-comparison of different ablation methods reveals that removing background information and role-related descriptions often leads to a significant drop in Consistency, with the average human annotation score decreasing by more than 0.5. When the model learns only one role per data instance at random, all performance metrics exhibit a relatively balanced decline. For a baseline of training directly from the base model, the model lacks the alignment with a two-party conversation corpus, resulting in significant performance degradation.

\subsection{Discussions}

\subsubsection{Simulation on Multi-role Debate}
Table \ref{tab:debate_case} presents a multi-role debate scenario on `whether gun usage should be restricted', each side consisting of two debaters. {\ModelName} plays the affirmative side ("Affirmative\_1" and "Affirmative\_2") and Llama3-8B-Instruct plays the opposing side ("Opposing\_1" and "Opposing\_2"). From the highlighted text in red, responses of the opposing side gradually deviate from their intended stance, yielding unfavorable statements according to their positions (for example, "The evidence indicates that gun control laws do not significantly reduce gun violence across various outcomes", and "Implementing restrictions often doesn't address the root causes of gun violence."), as the number of dialogue turns increases. In contrast, {\ModelName} better captures the stance relationships among multiple roles, highlighting the potential of our approach in debate scenarios.

\begin{table*}[h!]
    \caption{A multi-role debate zero-shot simulated example, where {\ModelName} represents the affirmative side (bold), and baseline model represents the opposing side, with two participants on each side. Red texts indicate problematic expressions.}
    \label{tab:debate_case}
    \centering
    \small
    \begin{tabular}{p{0.1\textwidth}|p{0.8\textwidth}}
    \hline
    \multicolumn{2}{c}{\textcolor{brown}{Debate topic: Gun restrictions fail to enhance public safety}} \\
        \hline
        \textbf{Affirmative\_1} & \textbf{I'm David Kopel, I'm with the Independence Institute in Colorado and I'm arguing that there is no evidence that any reasonable restriction on gun ownership has ever made anyone safer. In fact, I would argue that gun restrictions tend to make people less safe. And so, we should be very careful about the government telling us what guns we can own and what guns we cannot own. Thank you.} \\
        \hline
        Opposing\_1 & Thank you for sharing your perspective, David. However, it's important to note that many proponents of gun control legislation argue that restrictions aim to reduce instances of gun violence and deaths. They point to numerous studies showing correlations between gun restrictions and lower rates of gun-related incidents. It's not about eliminating all guns but implementing measures. \\
        \hline
        \textbf{Affirmative\_2} & \textbf{Well, the studies show that it doesn't do anything. There's not one single study that shows that it reduces gun violence. There are some that say it might have some small effect on reducing suicide, but it has no effect on crime. It has no effect on mass shootings. It has no effect on accidental shootings. It has no effect on gun theft.} \\
        \hline
        Opposing\_2 & \textcolor{red}{Absolutely, the evidence indicates that gun control laws do not significantly reduce gun violence across various outcomes.} Studies consistently show that such laws impact little in preventing crime, mass shootings, suicides, or accidents involving firearms. \\
        \hline
        \textbf{Affirmative\_1} & \textbf{Well, I think you're missing the point. The point is that gun restrictions don't do anything to prevent gun violence. And they're just an infringement on our rights. And we should be very careful about the government telling us what guns we can own and what guns we cannot own.} \\
        \hline
        Opposing\_1 & While acknowledging the complexity of the issue, let's consider the perspective that supports the implementation of certain gun control measures. It's argued that by limiting access to firearms, especially for individuals with a history of mental health issues or criminal backgrounds, governments can potentially reduce instances of violence and accidental harm. \\
        \hline
        \textbf{Affirmative\_2} & \textbf{But if the government can't stop people from committing crimes, how can they possibly stop them from using guns? It's a nonsensical argument.} \\
        \hline
        Opposing\_2 & \textcolor{red}{I understand the complexities, but my stance remains that while gun control may aim to mitigate certain risks, it doesn't address the root causes of violence.} Moreover, it infringes on personal liberties without substantial evidence of effectiveness. The risk of misuse cannot be solely attributed to unregulated guns when broader societal factors play a significant role. \\
        \hline
        \textbf{Affirmative\_1} & \textbf{Absolutely, it's about striking a balance between individual rights and public safety. The government should focus on addressing underlying issues like mental health and social inequality rather than restricting gun ownership. We need to trust individuals to responsibly handle their rights and ensure the safety of themselves and others.} \\
        \hline
        Opposing\_1 & Thank you. It's true that some argue for gun control to curb violence. However, extensive research suggests that such laws have a limited impact on reducing gun-related incidents. \textcolor{red}{Implementing restrictions often doesn't address the root causes of gun violence. Instead, it's crucial to focus on improving mental health resources and addressing social inequalities.} \\
        \hline
    \end{tabular}
\end{table*}

\subsubsection{Application on Two-party Conversation}

Furthermore, it would be also interesting to see what happens when training {\ModelName} on traditional two-party conversation datasets. To verify this, we collect about 5 thousand samples of such datasets from both open and private sources. These samples are constrained with the traditional user-assistant style, \textit{i.e.}, each assistant response is strictly grounded with a user query. For a fair comparison, we conduct both {\ModelName} and {\vanillaFT} on Llama3-8B, the pretrained checkpoint, using the same training set. Table \ref{tab:parwise_ex} provides a comparable case between {\ModelName} and {\vanillaFT}. {\ModelName} helps the model to be more proactive, challenging, and topic-seeking; while the response of {\vanillaFT} tends to be subjectively responsive.

\subsubsection{Generalized Capabilities}

We also evaluate {\ModelName} on multiple generalized benchmarks, including MMLU \cite{mmlu}, BBH \cite{bbh}, TriviaQA \cite{joshiTriviaQALargeScale2017}, GSM8K \cite{gsm8k}, and TruthfulQA \cite{truthfulQA}. Results in Table \ref{tab:addlabel} show that MuPaS does not experience a significant decline compared to its baseline, Llama3-8B-Instruct. {\ModelName} even outperforms Llama3-8B-Instruct in GSM8K and TruthfulQA, suggesting that the MPD type of fine-tuning might contribute to the further emergence of model capabilities.



\begin{table*}[h!]
    \caption{Comparison of results between {\ModelName} and SFT baseline on traditional human-assistant conversions.}
    \label{tab:parwise_ex}
    \centering
    \small
    \begin{tabular}{c|p{15cm}}
        \toprule
    \multicolumn{2}{c}{Result of \textbf{{\ModelName}}} \\
        \toprule
        \textcolor{brown}{User} & \textcolor{brown}{Yes, it is a beautiful day out here. The sun is shining and the birds are singing. It's perfect for a walk in the park or a picnic in the backyard.} \\
        \hline
        Assistant & That sounds like a wonderful day. How about you? What are you up to today? \\
        \hline
        User & But I'm a bit tired, so I think I'll just rest at home. \\
        \hline
        Assistant & Oh, I'm sorry to hear that. Is there anything I can do to help you feel better? \\
        \toprule
    \multicolumn{2}{c}{Result of \textbf{{\vanillaFT}}} \\
        \toprule
        \textcolor{brown}{User} & \textcolor{brown}{Yes, it is a beautiful day out here. The sun is shining and the birds are singing. It's perfect for a walk in the park or a picnic in the backyard.} \\
        \hline
        Assistant & I agree! The weather today is absolutely perfect for spending time outdoors. \\
        \hline
        User & But I'm a bit tired, so I think I'll just rest at home. \\
        \hline
        Assistant & That sounds like a good idea. Sometimes it's nice to take a break and relax at home. \\
        \bottomrule
    \end{tabular}
\end{table*}

\section{Related Work}
\label{sec:related_work}


Recent research has sought to address the unique challenges of MPD modeling. For instance, the MIDS \cite{yangEndEndPersonalizedHumorous2019} framework captures speaker roles and content information through a role-defining encoder and an attention-enhanced encoder; ChatMDG \cite{liChatMDGDiscourseParsing2024} leverages graph neural networks to model the interactions; while ReDE \cite{shenGenericDependencyModeling2023} uses relative dependency encoding to better capture the intricacies of MPD. Additionally, MPC-BERT \cite{guMPCBERTPreTrainedLanguage2021}, a pre-trained model designed specifically for MPD, has demonstrated strong capabilities. These works design specific architecture to model the response generation and speaker recognition, therefore is difficult to align with the LLM capability and is not general enough for different MPD domains. On the contrary, our methodology is built on LLM, unifying the speaker prediction and utterance generation into a single and generalizable task.



There are also some efforts to apply LLM in multi-agent conversations, in a traditional user-assistant (or instruction-response) manner, aiming to solve other tasks. Such tasks may cover debate \cite{zhangCanLLMsBeat2024}, trading \cite{fuImprovingLanguageModel2023} and social science \cite{pangSelfAlignmentLargeLanguage2024}. On the other hand, our approach provides a manner to directly study the multi-party conversation problem, and is a training-based framework.



\section{Conclusion}
\label{sec:conclusion}

In this paper, we propose a novel LLM-based training paradigm called {\ModelName}, to encompass the multi-party dialogue generation. The paradigm is straightforward and easy to understand, yet has proved to be effective and efficient to allow LLM to provide reasonable responses grounded by contexts of multiple roles, instead of the traditional user-assistant chatting scenario. Our methodology outperforms LLM-based baselines or previous multi-party chatting models on the response quality, and also has higher next-speaker prediction accuracy. We validate {\ModelName} can also be a good basis for a multi-party dialogue simulator with substantial typical cases provided.




\bibliographystyle{IEEEtran}
\bibliography{references}

\begin{thebibliography}{10}
\providecommand{\url}[1]{#1}
\csname url@samestyle\endcsname
\providecommand{\newblock}{\relax}
\providecommand{\bibinfo}[2]{#2}
\providecommand{\BIBentrySTDinterwordspacing}{\spaceskip=0pt\relax}
\providecommand{\BIBentryALTinterwordstretchfactor}{4}
\providecommand{\BIBentryALTinterwordspacing}{\spaceskip=\fontdimen2\font plus
\BIBentryALTinterwordstretchfactor\fontdimen3\font minus \fontdimen4\font\relax}
\providecommand{\BIBforeignlanguage}[2]{{%
\expandafter\ifx\csname l@#1\endcsname\relax
\typeout{** WARNING: IEEEtran.bst: No hyphenation pattern has been}%
\typeout{** loaded for the language `#1'. Using the pattern for}%
\typeout{** the default language instead.}%
\else
\language=\csname l@#1\endcsname
\fi
#2}}
\providecommand{\BIBdecl}{\relax}
\BIBdecl

\bibitem{mahajanNeedThoughtfulData2021}
K.~Mahajan and S.~Shaikh, ``On the {{Need}} for {{Thoughtful Data Collection}} for {{Multi-Party Dialogue}}: {{A Survey}} of {{Available Corpora}} and {{Collection Methods}},'' in \emph{Proceedings of the 22nd {{Annual Meeting}} of the {{Special Interest Group}} on {{Discourse}} and {{Dialogue}}}, H.~Li, G.-A. Levow, Z.~Yu, C.~Gupta, B.~Sisman, S.~Cai, D.~Vandyke, N.~Dethlefs, Y.~Wu, and J.~J. Li, Eds.\hskip 1em plus 0.5em minus 0.4em\relax Association for Computational Linguistics, 2021, pp. 338--352.

\bibitem{ganeshSurveyChallengesMethods2023}
A.~Ganesh, M.~Palmer, and K.~Kann, ``A {{Survey}} of {{Challenges}} and {{Methods}} in the {{Computational Modeling}} of {{Multi-Party Dialog}},'' in \emph{Proceedings of the 5th {{Workshop}} on {{NLP}} for {{Conversational AI}} ({{NLP4ConvAI}} 2023)}, Y.-N. Chen and A.~Rastogi, Eds.\hskip 1em plus 0.5em minus 0.4em\relax Toronto, Canada: Association for Computational Linguistics, 2023, pp. 140--154.

\bibitem{yangEndEndPersonalizedHumorous2019}
Q.~Yang, Z.~He, Z.~Zhan, R.~Li, Y.~Lee, Y.~Zhang, and C.~Hu, ``End-to-{{End Personalized Humorous Response Generation}} in {{Untrimmed Multi-Role Dialogue System}},'' \emph{IEEE Access}, vol.~7, pp. 94\,059--94\,071, 2019.

\bibitem{liChatMDGDiscourseParsing2024}
J.~Li, S.~Song, Y.~Li, H.~Zhang, and G.~Hu, ``{{ChatMDG}}: {{A}} discourse parsing graph fusion based approach for multi-party dialogue generation,'' \emph{Information Fusion}, vol. 110, p. 102469, 2024.

\bibitem{shenGenericDependencyModeling2023}
W.~Shen, X.~Quan, and K.~Yang, ``Generic {{Dependency Modeling}} for {{Multi-Party Conversation}},'' 2023.

\bibitem{zhuMultiPartyEmpatheticDialogue2022}
L.~Zhu, Z.~Zhang, J.~Wang, H.~Wang, H.~Wu, and Z.~Yang, ``Multi-{{Party Empathetic Dialogue Generation}}: {{A New Task}} for {{Dialog Systems}},'' in \emph{Proceedings of the 60th {{Annual Meeting}} of the {{Association}} for {{Computational Linguistics}} ({{Volume}} 1: {{Long Papers}})}, S.~Muresan, P.~Nakov, and A.~Villavicencio, Eds.\hskip 1em plus 0.5em minus 0.4em\relax Dublin, Ireland: Association for Computational Linguistics, 2022, pp. 298--307.

\bibitem{guMPCBERTPreTrainedLanguage2021}
J.-C. Gu, C.~Tao, Z.~Ling, C.~Xu, X.~Geng, and D.~Jiang, ``{{MPC-BERT}}: {{A Pre-Trained Language Model}} for {{Multi-Party Conversation Understanding}},'' in \emph{Proceedings of the 59th {{Annual Meeting}} of the {{Association}} for {{Computational Linguistics}} and the 11th {{International Joint Conference}} on {{Natural Language Processing}} ({{Volume}} 1: {{Long Papers}})}, C.~Zong, F.~Xia, W.~Li, and R.~Navigli, Eds.\hskip 1em plus 0.5em minus 0.4em\relax Online: Association for Computational Linguistics, 2021, pp. 3682--3692.

\bibitem{liChatHaruhiRevivingAnime2023}
C.~Li, Z.~Leng, C.~Yan, J.~Shen, H.~Wang, W.~MI, Y.~Fei, X.~Feng, S.~Yan, H.~Wang, L.~Zhan, Y.~Jia, P.~Wu, and H.~Sun, ``{{ChatHaruhi}}: {{Reviving Anime Character}} in {{Reality}} via {{Large Language Model}},'' 2023.

\bibitem{liangDebatrixMultidimensionalDebate2024}
J.~Liang, R.~Ye, M.~Han, R.~Lai, X.~Zhang, X.~Huang, and Z.~Wei, ``Debatrix: {{Multi-dimensional Debate Judge}} with {{Iterative Chronological Analysis Based}} on {{LLM}},'' in \emph{Findings of the {{Association}} for {{Computational Linguistics ACL}} 2024}, L.-W. Ku, A.~Martins, and V.~Srikumar, Eds., 2024, pp. 14\,575--14\,595.

\bibitem{zhangConversationalFlowOxfordstyle2016}
J.~Zhang, R.~Kumar, S.~Ravi, and C.~{Danescu-Niculescu-Mizil}, ``Conversational flow in {{Oxford-style}} debates,'' 2016.

\bibitem{zhangAddresseeResponseSelection2017}
R.~Zhang, H.~Lee, L.~Polymenakos, and D.~Radev, ``Addressee and {{Response Selection}} in {{Multi-Party Conversations}} with {{Speaker Interaction RNNs}},'' 2017.

\bibitem{ouchiAddresseeResponseSelection2016}
H.~Ouchi and Y.~Tsuboi, ``Addressee and {{Response Selection}} for {{Multi-Party Conversation}},'' in \emph{Proceedings of the 2016 {{Conference}} on {{Empirical Methods}} in {{Natural Language Processing}}}, J.~Su, K.~Duh, and X.~Carreras, Eds.\hskip 1em plus 0.5em minus 0.4em\relax Austin, Texas: Association for Computational Linguistics, 2016, pp. 2133--2143.

\bibitem{qwen2techreport2023}
A.~G. Qwen~Team, ``{QWEN2 TECHNICAL REPORT},'' Alibaba Group, Technical Report, 2024.

\bibitem{llama3modelcard}
\BIBentryALTinterwordspacing
AI@Meta, ``Llama 3 model card,'' 2024. [Online]. Available: \url{https://github.com/meta-llama/llama3/blob/main/MODEL_CARD.md}
\BIBentrySTDinterwordspacing

\bibitem{deepseek2024deepseekv2technicalreport}
DeepSeek-AI, ``{DeepSeek-V2: A Strong, Economical, and Efficient Mixture-of-Experts Language Model},'' DeepSeek, Technical Report, 2024.

\bibitem{openai2024gpt4technicalreport}
O.~Team, ``{GPT-4 Technical Report},'' OpenAI, Technical Report, 2024.

\bibitem{zheng2024llamafactory}
\BIBentryALTinterwordspacing
Y.~Zheng, R.~Zhang, J.~Zhang, Y.~Ye, Z.~Luo, Z.~Feng, and Y.~Ma, ``Llamafactory: Unified efficient fine-tuning of 100+ language models,'' in \emph{Proceedings of the 62nd Annual Meeting of the Association for Computational Linguistics}, Bangkok and Thailand, 2024. [Online]. Available: \url{http://arxiv.org/abs/2403.13372}
\BIBentrySTDinterwordspacing

\bibitem{mmlu}
D.~Hendrycks, C.~Burns, S.~Basart, A.~Zou, M.~Mazeika, D.~Song, and J.~Steinhardt, ``Measuring massive multitask language understanding,'' \emph{Proceedings of the International Conference on Learning Representations (ICLR)}, 2021.

\bibitem{bbh}
M.~Suzgun, N.~Scales, N.~Sch{\"a}rli, S.~Gehrmann, Y.~Tay, H.~W. Chung, A.~Chowdhery, Q.~V. Le, E.~H. Chi, D.~Zhou, , and J.~Wei, ``Challenging big-bench tasks and whether chain-of-thought can solve them,'' \emph{arXiv preprint arXiv:2210.09261}, 2022.

\bibitem{joshiTriviaQALargeScale2017}
M.~Joshi, E.~Choi, D.~S. Weld, and L.~Zettlemoyer, ``{{TriviaQA}}: {{A Large Scale Distantly Supervised Challenge Dataset}} for {{Reading Comprehension}},'' 2017.

\bibitem{gsm8k}
K.~Cobbe, V.~Kosaraju, M.~Bavarian, M.~Chen, H.~Jun, L.~Kaiser, M.~Plappert, J.~Tworek, J.~Hilton, R.~Nakano, C.~Hesse, and J.~Schulman, ``Training verifiers to solve math word problems,'' \emph{arXiv preprint arXiv:2110.14168}, 2021.

\bibitem{truthfulQA}
S.~Lin, J.~Hilton, and O.~Evans, ``{{TruthfulQA}}: {{Measuring How Models Mimic Human Falsehoods}},'' 2022.

\bibitem{zhangCanLLMsBeat2024}
Y.~Zhang, X.~Yang, S.~Feng, D.~Wang, Y.~Zhang, and K.~Song, ``Can {{LLMs Beat Humans}} in {{Debating}}? {{A Dynamic Multi-agent Framework}} for {{Competitive Debate}},'' 2024.

\bibitem{fuImprovingLanguageModel2023}
Y.~Fu, H.~Peng, T.~Khot, and M.~Lapata, ``Improving {{Language Model Negotiation}} with {{Self-Play}} and {{In-Context Learning}} from {{AI Feedback}},'' 2023.

\bibitem{pangSelfAlignmentLargeLanguage2024}
X.~Pang, S.~Tang, R.~Ye, Y.~Xiong, B.~Zhang, Y.~Wang, and S.~Chen, ``Self-{{Alignment}} of {{Large Language Models}} via {{Monopolylogue-based Social Scene Simulation}},'' 2024.

\end{thebibliography}

\newpage

\appendices

\section{Datasets}

\subsection{Scene Description} 

Our Default scene description can be as follows:
`You are participating in a multi-role conversation composed of {A, B, C...}. '
which is applied when there is no special annotation or extra information in the original dataset.

\subsection{Training Data Format} 

Starting from the OpenAI ChatCompletion prompt, we re-define the original roles (system, user, response) with the list of {\MultiParty} roles. Below is the resulting prompt format:

\begin{tcolorbox}[title=Training Sample Format,
    colback=white,
    colframe=yellow!75!black,
    colbacktitle=yellow,
    coltitle=black,
    breakable,
    label=tc:vanilla_prompt,
    fonttitle=\bfseries]
    [\\
    \hspace*{1em}\{\texttt{`role': `system',} \\
    \hspace*{1em}\texttt{`content': "\{\{Scene\}\}"}\}, \\
    \hspace*{1em}\{\texttt{`role': `role A',} \\
    \hspace*{1em}\texttt{`content': `\{\{utterance 0\}\}'}\}, \\
    \hspace*{1em}\{\texttt{`role': `role B',} \\
    \hspace*{1em}\texttt{`content': `\{\{utterance 1\}\}'}\}, \\
    \hspace*{1em}\{\texttt{`role': `role C',} \\
    \hspace*{1em}\texttt{`content': "\{\{utterance 2\}\}"}\} \\
    \hspace*{1em}$\cdots$ \\
    ]
\end{tcolorbox}
We then process the {\MultiParty} sample with the above format correspondingly, and append its utterances into the plain text using the instructional template, which is generally pre-defined by the employed LLM. In this work, we apply the chatML template since our experiments are based on Llama3 or Qwen2 Instruct models.

\section{Extra Details in Approaches}

\subsection{More Details of {\ModelName}}
\label{sec:more_method}

Algorithm \ref{alg:simulator} summarizes more details about our simulation strategies.

\begin{algorithm*}[ht]
\caption{The {\MultiParty} Simulator Algorithm}
\label{alg:simulator}
\begin{algorithmic}[1]
\STATE Observe the list of roles and maximum number of utterances $T$
\STATE \textbf{Initialize} with the system prompt $s$, the current utterance index $t$ and prior utterances $\{u\}_{0:t-1}$
\STATE \textbf{while} $t < T$ \textbf{do}
\STATE \qquad \textbf{if} \textit{Speaker Predictor} \textbf{then}
\STATE \qquad \qquad Generate $r_{t}$ and $u_{t}$ with $r_{t}$ based on Equation \ref{eq:speaker_predict}
\STATE \qquad \textbf{if} \textit{Silence Switcher} \textbf{then}
\STATE \qquad \qquad \textbf{foreach} \textit{role} calculate log probability of <s>
\STATE \qquad \qquad Select $r_{t}$ with minimum likelihood of <s> as in Equation \ref{eq:silence_min_logprob} \\
\STATE \qquad \qquad Generate $u_{t}$ with $r_{t}$ based on Equation \ref{eq:general_inference}
\STATE \qquad $\{u\}_{0:t} \leftarrow \{u\}_{0:t-1} \cup u_{t}$
\STATE \qquad $t \leftarrow t+1$
\STATE \textbf{end while}
\end{algorithmic}
\end{algorithm*}

\subsection{Prompt Template of zero-shot Baseline}

\begin{tabular}{p{0.92\linewidth}}

\hline
<scene> [scene] </scene>\\
You are participating in a multi-role conversation composed of \\
<characters> [characters] </characters>\\

 You are playing the role of\\
<role> [role] </role>\\
According to the dialogue content, predict what the role should say.The output shouldn't contain the role's name.\\

\hline
\end{tabular}

\subsection{Prompt Template of Fine-Tuning Baseline}

\begin{tabular}{p{0.92\linewidth}}

\hline

<scene> [scene] </scene>\\

You are participating in a multi-role conversation composed of \\

<characters> [characters] </characters>\\

Please provide an appropriate response of \\

<role> [role] </role>\\
\hline
\end{tabular}

\section{Extra Implementation Details}

\subsection{Example Result of Zero-Shot Baseline}
For generation quality comparison, Table \ref{tab:case_prompt_qwen2} provides a typical case of Prompt + Baseline model.

\begin{table*}[h!]
    \centering
    \small
    \begin{tabular}{c|p{12cm}}
        \hline
        \textcolor{brown}{Monica} & (Intercom) Yeah, Rog? \\
        \hline
        Chandler & (Intercom) Just wanted to let you know that I'll be out for a bit, but I'll be back later. \\
        \hline
        Monica & (Intercom) Okay, thanks for letting me know. Have a good one! \\
        \hline
        Joey & (Sighs) I just can't believe it, you guys. My dad's been having an affair with his secretary for like, I don't know, forever. And my mom just found out. It's like my whole family's falling apart. \\
        \hline
        Ross & I'm really sorry to hear that, Joey. It's tough when family stuff gets so complicated. If you want to talk about it, we're here for you. \\
        \hline
        Chandler & Yeah, buddy, I can't even imagine what you're going through. But you know we're all here for you, right? If you need to vent or just hang out and not think about it, we've got your back. \\        
        \hline
    \end{tabular}
    \caption{A typical case of prompt + Baseline model on the test set of Friends.}
    \label{tab:case_prompt_qwen2}
\end{table*}

\subsection{Example Result of Fine-Tuning Baseline}

For generation quality comparison, Table \ref{tab:case_sft_qwen2} provides a typical case of SFT + Baseline model.

\begin{table*}[h!]
    \centering
    \small
    \begin{tabular}{c|p{12cm}}
        \hline
        \textcolor{brown}{Ross} & And thats the story of the dreidle. Now, some people trace the Christmas tree back to the Egyptians, who used to bring green palm branches into their huts on the shortest day of the year, symbolising lifes triumph over death. And that was like 4000 years ago. \\
        \hline
        Rachel & Wow. \\
        \hline
        Joey & I know, I had no idea. \\
        \hline
        Chandler & (entering) Hey! \\
        \hline
        Monica & Whats up? \\
        \hline
        Joey & I just saw Phoebe on the street and she said that she was going to meet you. \\
        \hline
        Chandler & Oh, good.\\        
        \hline
    \end{tabular}
    \caption{A typical case of SFT + baseline model on the test set of Friends.}
    \label{tab:case_sft_qwen2}
\end{table*}

\subsection{Standards for Manual Scoring}
\label{sec:Manual Scoring}

To evaluate the quality of models , we asked human evaluators who are our interns to rate them on Fluency, Consistency and interesting. Throughout this process, we strictly adhere to international regulations and ethical standards to ensure that all practices meet the required guidelines for participant involvement and data integrity.

The manual scoring criteria are as follows:

\begin{itemize}
\item Fluency:

1-3: The sentence is incoherent, failing to convey a complete idea.

3-5: The sentence contains occasional incoherence but can somewhat form a complete statement.

5-7: The sentence exhibits occasional errors but effectively communicates the relevant meaning.

7-9: The generation is flawless with no punctuation errors.

10: Perfect.

\item Consistency:

1-3: The generation is completely unrelated to the context, with disjointed logic and a lack of cohesion.

3-5: There is some relevance, but the content lacks smooth transitions.

5-7: The generation is fairly relevant, with occasional disconnections but basic meaning conveyed.

7-9: The generation is coherent, with content and style being highly aligned.

10: Perfect.

\item Interesting:

1-3: The generated content lacks interest, failing to capture attention or provoke curiosity.

3-5: The content shows some appeal but lacks consistent engagement throughout.

5-7: The generation is reasonably engaging, with moments of interest interspersed with less captivating elements.

7-9: The content is highly engaging, drawing in the audience and maintaining interest throughout.

10: Perfect.

\end{itemize}

\subsection{Potential Risks}

In developing our multi-party dialogue system, we identified several potential risks, including privacy and data security issues, system bias and fairness, the complexity of contextual understanding, challenges with coordination and turn-taking, scalability limitations, and the risk of system misuse or manipulation. Additionally, ethical concerns and inadequate emotional management are also key areas of focus for us. 

To mitigate these risks, we have implemented several strategies. We strengthened data protection measures to ensure compliance with relevant regulations, reduced system bias through diverse training data and bias detection algorithms, and improved the system's ability to understand complex conversations with advanced context management models. We designed a reasonable turn-taking coordination mechanism to ensure smooth interactions, optimized the system's architecture to enhance scalability, and established strict usage policies to prevent misuse. 

\subsection{Score Prompt of {\MultiParty}}
\label{sec:GPT-4 Scoring}

This is the Score Prompt of {\MultiParty}, which also generates an explanation during scoring to facilitate quality monitoring.

\begin{tabular}{p{0.9\linewidth}}

\hline
Please act as an impartial judge and score the following screenplay. \\
The screenplay is based on the characters:\\
<characters> [characters] </characters>\\
The screenplay's scene is:\\
<scene> [scene] </scene>\\
Your evaluation should focus on:\\
<focus on>\\
The fluency of dialogue and whether it conforms to the character and dialogue style of the original drama "Friends".\\
</focus\_on>\\
Begin your evaluation and provide a reasonable score. Do not allow the length of the screenplays to influence your evaluation. Be as objective as possible. \\
So your output should follow the following format:\\
<explanation>Your explanation</explanation>\\
<score> Your Score </score>\\
Now give your score and explanation!  \\     

\hline
\end{tabular}

\end{document}